\documentclass{article}

\usepackage{arxiv}

\usepackage[utf8]{inputenc} 
\usepackage[T1]{fontenc}    
\usepackage{hyperref}       
\usepackage{url}            
\usepackage{booktabs}       
\usepackage{amsfonts}       
\usepackage{nicefrac}       
\usepackage{microtype}      
\usepackage{lipsum}
\usepackage{graphicx}
\usepackage{float}
\usepackage{amsmath}

\usepackage{orcidlink}

\title{IMUDiffusion: A Diffusion Model for Multivariate Time Series Synthetisation for Inertial Motion Capturing Systems}

\author{
 Heiko Oppel \orcidlink{0000-0002-4410-2442} \\
  Research Group Biomechatronics\\
  Ulm University of Applied Sciences\\
  89081 Ulm, Germany \\
  \texttt{heiko.oppel@thu.de} \\
   \And
 Michael Munz \orcidlink{0000-0003-3427-3827} \\
  Research Group Biomechatronics\\
  Ulm University of Applied Sciences\\
  89081 Ulm, Germany \\
  \texttt{michael.munz@thu.de} \\
}

\begin{document}
\maketitle
\begin{abstract}
Kinematic sensors are often used to analyze movement behaviors in sports and daily activities due to their ease of use and lack of spatial restrictions, unlike video-based motion capturing systems. 
Still, the generation, and especially the labeling of motion data for specific activities can be time-consuming and costly. 
Additionally, many models struggle with limited data, which limits their performance in recognizing complex movement patterns. 
To address those issues, generating synthetic data can help expand the diversity and variability.
In this work, we propose IMUDiffusion, a probabilistic diffusion model specifically designed for multivariate time series generation. 
Our approach enables the generation of high-quality time series sequences which accurately capture the dynamics of human activities. 
Moreover, by joining our dataset with synthetic data, we achieve a significant improvement in the performance of our baseline human activity classifier. 
In some cases, we are able to improve the macro F1-score by almost $30\%$.
IMUDiffusion provides a valuable tool for generating realistic human activity movements and enhance the robustness of models in scenarios with limited training data.
\end{abstract}


\section{Introduction}
Human activity recognition (HAR) plays a critical role in numerous applications, including healthcare and sports monitoring and smart environments. 
The accurate recognition of human activities from sensor-based time series data is an ongoing challenge due to the complexity and variability of human motions. 
However, signal analysis and deep learning techniques, have significantly advanced the field. 
Particularly, generative models have shown promising results in generating high-quality data across various domains, including images and time series. \newline
\newline
The synthesis of realistic time series data for human activities is essential for multiple reasons, such as increasing limited datasets, reducing the need of additional labeling and improving model robustness. 
Up to date, there exist several types of generative models, such as Generative Adversarial Networks (GANs) \cite{iniGAN}, \cite{LSTMCNN_GAN}, \cite{GANBCI}, \cite{GANonHAR} and Variational Autoencoders (VAEs) \cite{vae_multivariat}, which have been applied to time series generation. 
However, they often struggle with mode-collapse \cite{modecollapse} and variability in data length \cite{GANLengthProblem}. 
In contrast, Denoising Diffusion Probabilistic Models (DDPMs) may offer a more versatile approach by learning a reverse diffusion process to receive the synthetic signals.\newline
\newline
In this work, we propose a novel framework IMUDiffusion for synthesising human activity time series data using denoising diffusion probabilistic models. 
The data were recorded using inertial measurement unit (IMU) sensor data.
Such models are already widely used in the domain of Computer Vision \cite{ini_DDPM} and Natural Language Processing, especially, the combination of both domains \cite{nlpAndImgDDPM}.
In recent years, the intereset in biophysiological data generation increased, which also brought models like DDPM into focus.
Several work groups already provided a comprehensive survey on diffusion models and their application in general \cite{ddpmSurvey_1}, \cite{ddpmSurvey_2}.
Comparable works with DDPM on time series have been provided by Kont et al. \cite{kong2021diffwave} where they developed a diffusion model for conditional and unconditional waveform generation or Tashiro et al. \cite{imputationDDPM} and Alcaraz et al. \cite{imputationForecastDDPM} as they investivated into time series imputation and forecsating. \newline
DDPMs have also already been developed for kinematic data.
Li et al. \cite{DDPMIMU} developed a DDPM for biomedical signals.
They tested their model on three different datasets.
The first one is a synthetic dataset based on pre-defined signal patterns, whereas the two other datasets are real world sets. 
One is MIT-BIH Arrhythmia dataset \cite{mit-bih} containing electrocardiogram signals and the other one is the UniMiB dataset \cite{unimib}.
For the UniMiB dataset participants wore a smartwatch from which they recorded accelerations in three dimensions.
Their UNet architecture is build around a down-, a mid- and an up-block.
As input receives the network the time series data in addition to the time step embedding and other conditional embeddings. \newline
In this work, we developed a model to fit the needs of generating real world kinematik data resembling data from a six axis IMU.
The architecture is based on the UNET architecture for image generation \cite{ini_DDPM}.
We evaluated our IMUDiffusion model by analysing visual similarities and by testing it against a baseline classifier which distinguishes several human activities. \newline

This work has the following main contributions:
\begin{itemize}
    \item We propose a diffusion model for time series synthesis for IMU based motion capturing systems utilizing the frequency domain.
    \item We analyse the diffusion process of the signal and evaluate the quality of the signals in comparison with real-world data.
\end{itemize}

\section{Methods}
\subsection{Human Activity Recognition Dataset}
To evaluate the model performance, we used the benchmark dataset from Banos et al. \cite{displacement_dataset} where they analysed sensor displacement in human activity recognition tasks. 
This dataset consists of 33 human activities which were recorded with 17 participants. 
Each participant was wearing nine inertial measurement units from Movella.
One of their aims was to analyse the impact of displaced inertial sensors.
Therefore, they recorded the same activities with all participants in three different setups. 
One was an ideal placement of the sensors attached by the researcher, the second setup was about self-placement of the sensors and the last one about mutually displacement. 
For our study, we used the recordings from the ideal placement setup to guarantee most similar IMU positions across all participants.
For an easier interpretation of the synthetic data, we reduced the amount of IMUs to a single IMU located on the right thigh. 
This position contains enough information over all chosen classes.
From the 33 classes, we selected four activities to reduce the necessary computation time for the synthetisation process and classification.
The four chosen activities are Walking, Running, Jump Up and Cycling.
Each participant had to perform the activities of Running, Walking and Cycling for one minute.
The Jump Up activity on the other side was recorded 20 times. 
So, each participant did jump up 20 times in a row which was recorded by the IMUs.
The classes were specifically selected to address several real-world challenges.
One of them is a class imbalance in the dataset where one class is underrepresented (minority class problem). 
In our case, this is the Jump Up class.
Another challenge is a high resemblance of samples between classes.
Walking and Running fulfill this criteria as the execution of those movements are similar.
The aim of our study is to differentiate those activities using a deep neural network.
In combination with the synthetic data, we aim to further improve the differentiability. \newline
To evaluate such a classifier, a hold out test set is required. 
We were using the leave one subject out cross validation (LOSOCV) principle, where we trained our models on $n-1$ participants and tested it against the remaining participant not included in the training process.

\subsection{Signal Processing}\label{section:signal_processing}
For some participants their activities are missing.
This is mostly the case for the Cycling class.
Those participants were neglected in this study as we require all of the four activities for the classification task.
It reduced the pool of participants to $12$.
The remaining participant IDs (PIDs) are: $1, 2, 3, 5, 8, 9 ,10, 11, 12, 13, 14$ and $16$. \newline

\subsubsection{Recording Information}
The four classes vary in their length of recording. 
Walking, Running and Cycling were supposed to be recorded for a complete minute.
Though, they vary from $44.64\,s$ to $103.02\,s$.
The class Jump Up on the other side is the minority class in the dataset and contains recording times from $9.34\,s$ up to $12.44\,s$.
The whole variation can be seen in Table~\ref{tab:recording_times}.

\begin{table}[h]
    \centering
    \caption{Time duration of the recording of the individual activities.}
    \begin{tabular}{|c|c|c|c|}
        \hline
        Activity & Average Time [s] & Minimum Time [s] & Maximum Time[s]\\
        \hline
        Walking & $72.55 \pm 16.42$ & $52.18$ & $100.28$ \\
        Running & $52.59 \pm 5.03$ & $44.64$ & $62.44$ \\
        Jump Up & $11.12 \pm 0.91$ & $9.34$ & $12.44$ \\
        Cycling & $69.57 \pm 15.66$ & $46.58$ & $103.02$\\
        \hline
    \end{tabular}
    \label{tab:recording_times}
\end{table}

\subsubsection{Signal Sequencing}\label{section:signal_sequencing}
\begin{figure}[H]
    \centering
    \includegraphics[width=0.75\linewidth]{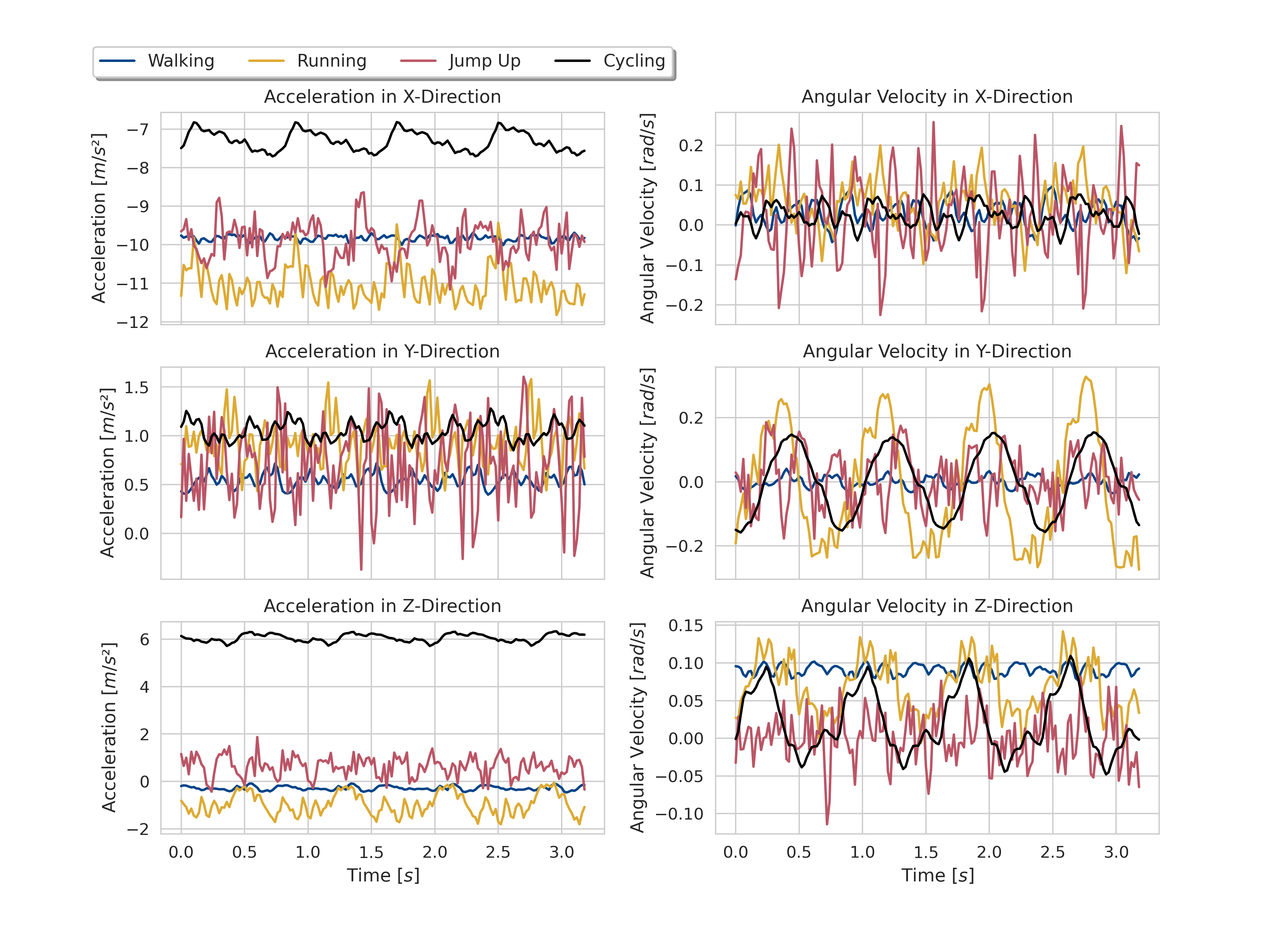}
    \caption{Sequenced sensor information in their respective SI-units separated by activity and sensor axis. The sequences were cut to $160$ timesteps with a shift of $40$ steps. Each activity contains at least three repetitive movements of the same activity. The graph visualizes the average over all participants.}
    \label{fig:raw_sensor_information.}
\end{figure}

In order to synthesise human activities recorded with inertial sensors, a certain amount of training samples are required. 
Therefore, the recordings of the participants had to be cut to shorter time sequences to increase the sample space. 
In combination with a time sequence shift, the sample space was even further increased.
The activities were recorded with a sampling rate of $50\,Hz$, leading to a number of samples from $467$ (Jump Up class) to $5151$ (Cycling class) per recording.
With the aim of synthesising multiple repetitions of a movement at once whilst not reducing the minority class too far, we decided for a sequence size of $160$ timesteps with a shift of $40$ steps. 
This means, we have an overlap of $120$ timesteps between two sequences, of $80$ timesteps between three and of $40$ timesteps between four sequences.
\newline
Figure~\ref{fig:raw_sensor_information.} visualizes the average of the four activities individually for each sensor axis of the sequenced recordings.
Each sequenced activity stores information of at least three repetitive movements. 
Most of those movements took place in the sagittal plane.
Naturally, this reduced the amount of movement information in the coronal plane.
Based on the orientation of the IMU, this reduced the information content of the z-axis of the accelerometer, and, concurrently the x-axis of the gyroscope for the purpose of activity classification. \newline
After the sequencing of the signals, we standardized the signals further.
Each axis of the IMU was standardized separately. 

\subsubsection{Signal Transformation}
In this study, we examined the possibility of synthesising time series sequences of human movements. 
Therefore, we transformed the initial signal to the frequency domain by computing the short time fourier transform (STFT).
We used a hanning window with a length of 22 steps and an overlap of 20. 
Due to the resulting resolution, this reduced the frequency dimension to $12$ and the time dimension to $80$ steps.
Another dimension is made up of the $12$ channel
They are made up of two times the six IMU axis due to the real and imaginary parts after the STFT.

\subsection{Denoising Diffusion Probabilistic Model}
The input data of our IMUDiffusion model is the scaled transformation of the time series sequences in the frequency domain as described in Section~\ref{section:signal_processing}.
The architecture of the IMUDiffusion was adapted from the image models to generate human movement sequences from white gaussian noise. 

\subsubsection{The IMUDiffusion Architecture}
\begin{figure}[H]
    \centering
    \includegraphics[width=0.75\linewidth]{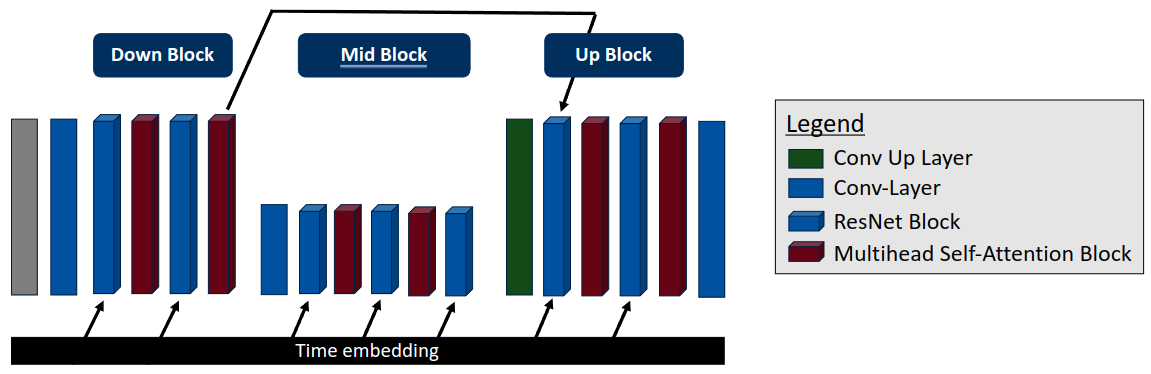}
    \caption{
    The network architecture of the IMUDiffusion model. 
    It consists of three blocks: a Down, Mid and Up block. 
    Each block is build around two ResNet and two Multihead Self-Attention blocks connected in serial.
    }
    \label{fig:diffusion_network}
\end{figure}

The Network architecture of the IMUDiffusion can be seen in Figure~\ref{fig:diffusion_network}.
The basis of the architecture is build on the basis of Jonathan Ho et al.~\cite{ini_DDPM} diffusion model which established ResNet and Attention layer in diffusion models.
In the same way, we focused on the typical three block architecture with skip connections. 
The base channel were reduced to 32 and only a single dimension reduction was applied along the time domain alone.
The kernel dimension was defined analogously.
It only convolves across the time domain. 
For the time embedding, we rely on the sinusoidal time embedding as stated in Equation~\ref{eq:time_embedding}.
The dimension $t_{dim}$ was set to $128$.
The variable $t$ defines the current time step.
For more information on the DDPM architecture see \cite{ini_DDPM}.

\begin{equation}
    TE_t = [\sin{(\frac{t}{10000^{i/{t_{dim}}}})}, \cos{(\frac{t}{10000^{i/{t_{dim}}}})}] \text{, with } i\in[0, t_{dim}] \label{eq:time_embedding}
\end{equation}

\subsubsection{Forward Diffusion Process}
In order to add noise to the initial time series, we used a linear scheduler.
It regulates the amount of noise added at time step $t$.
Typically, the process of adding noise is iterative, where at each step, the sequence gradually approximates the characteristics of gaussian white noise.
The number of timesteps to add noise to the initial input is decisive for the outcome.
The initial paper from Jonathan Ho et al.~\cite{ini_DDPM} used $1000$ timesteps.
In our case, this was not enough to guarantee that a smooth enough transition from a noisy sequence to an expressive human motion was guaranteed. 
Therefore, we had to increase the number of timesteps to $T=3000$.
Another challenge was the identification of the optimal $\beta$ value, as each sensor reacts differently on the scheduler.
Therefore, we separated the linear schedulers diffusion rate $\beta$ to match them individually.
The chosen values are $\beta_{Acc}=9e{-4}$ and $\beta_{Gyro}=6e{-4}$.


\subsubsection{Training of the IMUDiffusion model}
The IMUDiffusion was trained for $4500$ epochs.
An Adam optimizer with a learning rate of $\eta=4e{-4}$ was used in combination with a smooth L1 loss with a beta value of $\beta_{L1}=1.0$.
Throughout each step within an epoch random noise was added to the transformed time series. 
The added noise depends on the scheduler, the diffusion rate $\beta$ and the sequence at time $t-1$ and is described by Equation~\ref{eq:noisy_sample} \cite{ini_DDPM}.

\begin{equation}
    x_{k,t} = \sqrt{1-\beta_{k,t}}x_{k, t-1} + \sqrt{\beta_{k, t}}\epsilon \text{, with } \epsilon \sim \mathcal{N}(0, I) \text{, } t \in \{\mathbb{N} | 0<t<3000\} \text{, } k\in\{Acc, Gyro\}\label{eq:noisy_sample}
\end{equation}

The duration of the training using DDPM for synthetisation depends on the size of the dataset and the hardware.
As the dataset consisted of $22$ sequences per class and participant, the training of the IMUDiffusion model required approximately $8$ minutes each. 
The synthetisation process for a batch of $128$ sequences required another $3$ minutes per participant and class.
As we evaluated $12$ participants, the training time summed up to $78.4$ computing hours.
With $72$ computing hours, most of the time was spent on the synthetisation instead of the training of the model.
The calculations were performed on an NVIDIA RTX 3090. 
Overall, we generated $3840$ sequences for each class and LOSOCV step.

\subsection{Classification}
In the LOSOCV classification task we initially started without any synthetic sequence, meaning, the model is trained only on real sequences and tested on real sequences.
Then, we added $1\%$ of synthetic sequences and started to train the network anew. 
This was repeated until all synthetic sequences were used for training. \newline
First, the network architecture is defined including the hyperparameter initialization, before the different model are explained.

\subsubsection{Classifier Model}
The classifier was designed to differentiate between the four classes chosen from the human activity dataset, namely, Walking, Running, Jump Up and Cycling.
It is build around a Convolutional Neural Network consisting of three convolutional layer for feature extraction followed by three linear layers.
Each convolutional layer convoles the input only along the time dimension with a kernel of size $5\times1$.
A MaxPooling layer is used to halve the dimension of the time after the second convolutional layer.
The first two dense layers are followed by a dropout layer with a $p-$value of $0.3$.
Additionally, the classifier block uses an L2-regularization with $\lambda=1e-4$.

\subsubsection{Model Definition}
The 2 Sample Full Synth classifier is compared against two different baseline classifier. 
The first one only uses the real training sequences on which the IMUDiffusion model was trained on. 
This leads to a reduced sample space of $88$ sequences in the training set per LOSOCV step.
For the validation set, we randomly chose $22$ sequences per class from the same participants not included in the test set.
This first baseline model is further referenced as the $2$ Sample classifier.
Though, for each participant around $2600$ to $2700$ sequences are available for training.
So, for the second baseline classifier, we used $80\,\%$ of the available data for training.
Still, with one hold out test subject.
The remaining $20\,\%$ were used to monitor the training process.
This second baseline classifier is further referenced as Full-Set.

\subsection{Methods to Analyse Synthetic Sequences}
We chose two approaches to analyse the similarity between real and synthetic sequences. 
The first one is a dimensionality reduction technique and the second one uses a cluster approach. 

\subsubsection{UMAP for Visualization}
Uniform Manifold Approximation and Projection (UMAP) \cite{UMAP} is an approach to reduce the dimensionality within a data set.
It is a widely used approach to visualize the similarity between synthetic and real sequences \cite{DDPMIMU}, \cite{umap_2}, \cite{umap_3}, \cite{umap_4}.
It is able to reduce the feature space to two dimensions, enabling a visual feedback about the feature space.

\subsubsection{kMeans on Time Series for Visualization}
Due to the challenges of interpreting the results obtained from the UMAP algorithm, we used a second approach to visualize and compare the synthetic sequences against the real ones.
We clustered the sequences per sensor axis using kMeans clustering with dynamic time warping (DTW) as distance metric to account for the temporal offset.
The number of cluster was set to $20$ for the $3840$ samples.
We also included the DTW barycenter average (DBA) \cite{DTWBarycenterAveraging} representing the cluser center.
The DBA converges over all relevant sequences taking the temporal offset into account, which makes it easier to interpret and compare the real against the synthetic sequences.


\section{Results}
The results section is divided in three parts. 
The first part presents a visual analysis of the synthetic sequences.
The second part examines the denoising process using a synthetic sequence as an example.
Finally, the third part analyses the classification result.

\subsection{Synthetic Signal Evaluation}
\begin{figure}[H]
    \centering
    \includegraphics[width=0.99\linewidth]{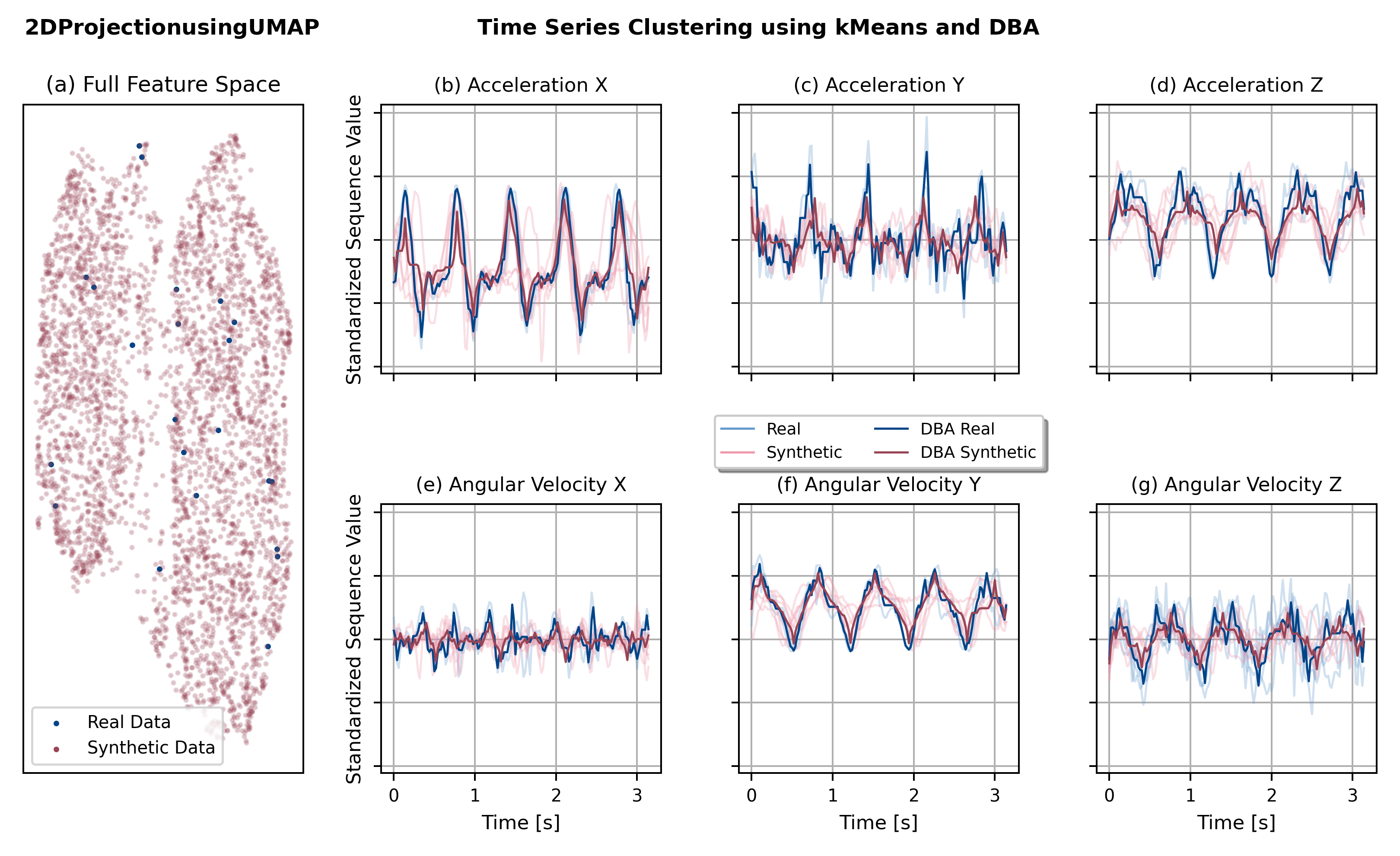}
    \caption{
    Comparison of synthetic and real sequences from the Cycling class using UMAP (a).
    Two cluster are formed by the synthetic sequences whereas the real sequences spread individually across the dimension space.
    The subgraphs (b)-(g) visualize one of the $20$ clusters from the Cycling class using kMeans clustering for time series including the DTW barycenter average for the real and synthetic sequences separately.
    Each subgraph visualizes eight randomly chosen sequences belonging to the pre-chosen cluster.
    They also represent one of the six axes of the IMU separately.
    The bold lines belong to the DTW barycenter average of the real and synthetic sequences respectively.
    }
    \label{fig:seq_analysis}
\end{figure}

In this section, the synthetic sequences are visually analysed.
Therefore, we used the results from the Cycling class throughout the sections.
It is the class which is easiest to interpret visually, as most of the sensor axes contain movement information.
Figure~\ref{fig:seq_analysis} visualizes the results from the UMAP dimensionality reduction (a) and the kMeans clustering approach (b)-(g) for this class.

\subsubsection{UMAP for Visualization}
UMAP allows to reduce the dimensionality of a high dimensional dataset to evaluate similarities visually.
Reducing the dimension of the synthetic sequences to two, generated two main cluster.
This could be seen within each class and is exemplary visualized in Figure~\ref{fig:seq_analysis} (a) for the Cycling class. 
The real sequences on the other hand distributed over a broader space with multiple isolated sequences and small cluster.
Though, they are still bound by the two cluster from the synthetic data.
A possible explanation is the noisy nature of the IMU data.
Additionally, the sequencing process of the signal, see Section~\ref{section:signal_processing} is responsible for small shifts within the signal which is difficult to address using UMAP. \newline
Even by separating the feature space into the individual sensor axes did not affect the result.
The real sequences did not form a cluster considering multiple sequences.
They remained mostly isolated, but were still located within the cluster from the synthetic sequences.

\subsubsection{kMeans on Time Series for Visualization}
Exemplarily Figure~\ref{fig:seq_analysis} (b)-(g) visualize eight real and eight synthetic sequences for one specifically selected cluster.
Based on the visual interpretation of the clustering results, our IMUDiffusion model was able to generate sequences with similar characteristics to the real sequences.
For example, the angular velocity in x-direction (Figure~\ref{fig:seq_analysis} (e)) contains the least movement information due to the placement of the IMU, see Section~\ref{section:signal_sequencing}.
Still, our model generated similar patterns matching the ones from the real sequences.

\subsection{Evaluation of the Denoising Process}
\begin{figure}[H]
    \centering
    \includegraphics[width=0.99\linewidth]{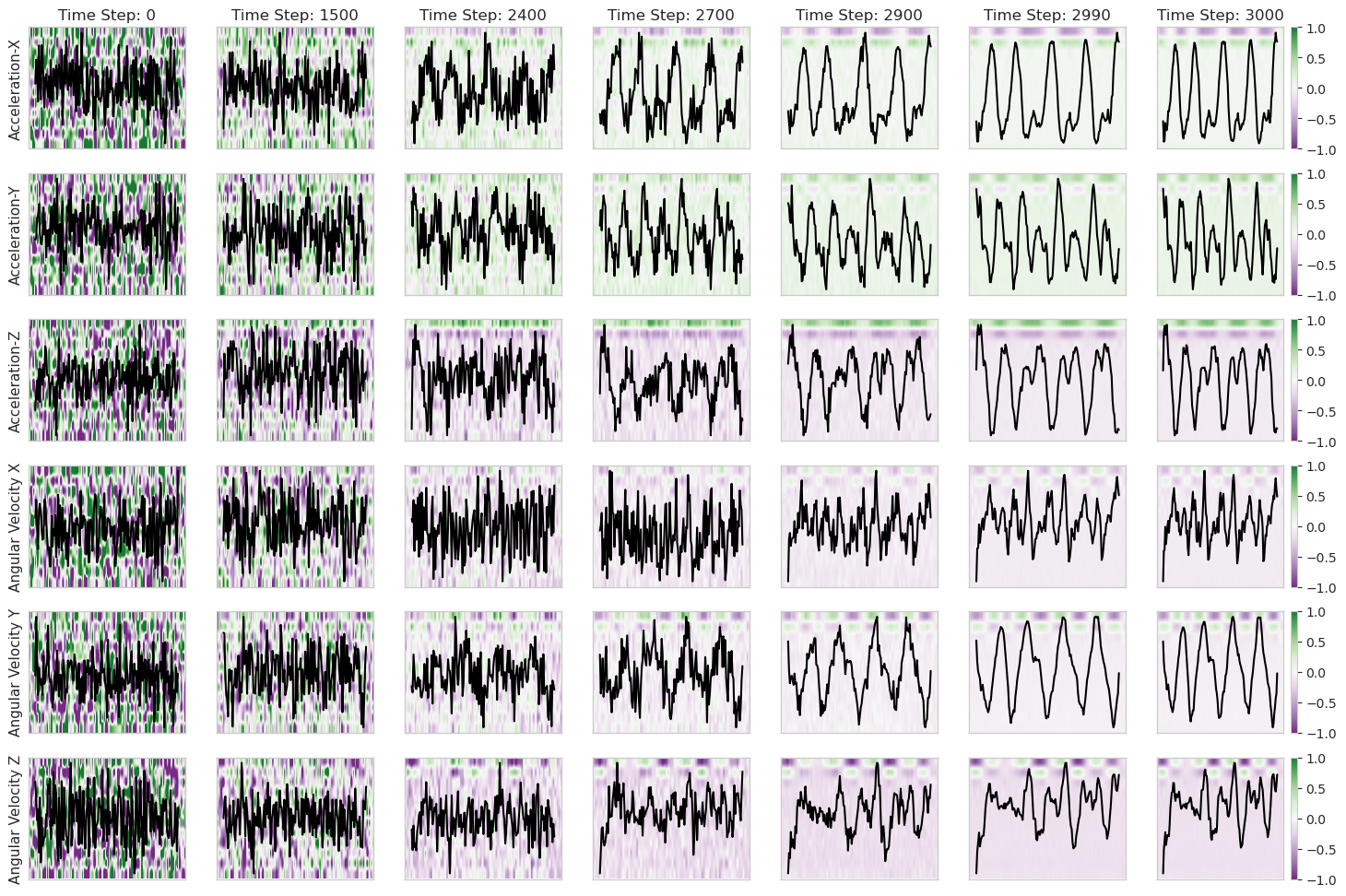}
    \caption{
    Visualization of the denoising process on an example from the Cycling class.
    The background within the images show the transformed sequence into the frequency domain.
    The black lines visualize the corresponding sequence in the time domain.
    The denoising process is separated by the IMU axes.
    }
    \label{fig:denoising_example}
\end{figure}

The following examination of the denoising process is a visual analysis including a subjective interpretation of the results.
It is based on a synthetic example from the Cycling class which is visualized in Figure~\ref{fig:denoising_example}.
There, the denoising process for each of the six IMU axes is separated to demonstrate the cross-sensor effectiveness of our approach.
Beginning from the initial time step zero, we visualized six additional time steps, including the last one at time step $3000$.
In some cases, the movement characteristics begin to develop approximately $2400$ steps into the denoising process, for example in case of the acceleration in x-direction.
Though, the high frequency noise within the sequences have to be further removed to generate comparable results to the real sequences.
Around the $2900^{th}$ step, the characteristics are getting increasingly pronounced.
Still, the amount of noise further reduces until the final step.
Though, already before the last step, around the $2990^{th}$ step, the signals characteristics are clearly visible and difficult to distinguish from the real sequences. \newline
Even though, the angular velocity in x-direction contains the least information regarding the human activity, the IMUDiffusion model learns some characteristics from the real sequences it was trained on.
As it could be seen in Figure~\ref{fig:seq_analysis} (e) those patterns match with the real sequences.

\subsection{Classification Results}
\begin{figure}[H]
    \centering
    \includegraphics[width=0.99\linewidth]{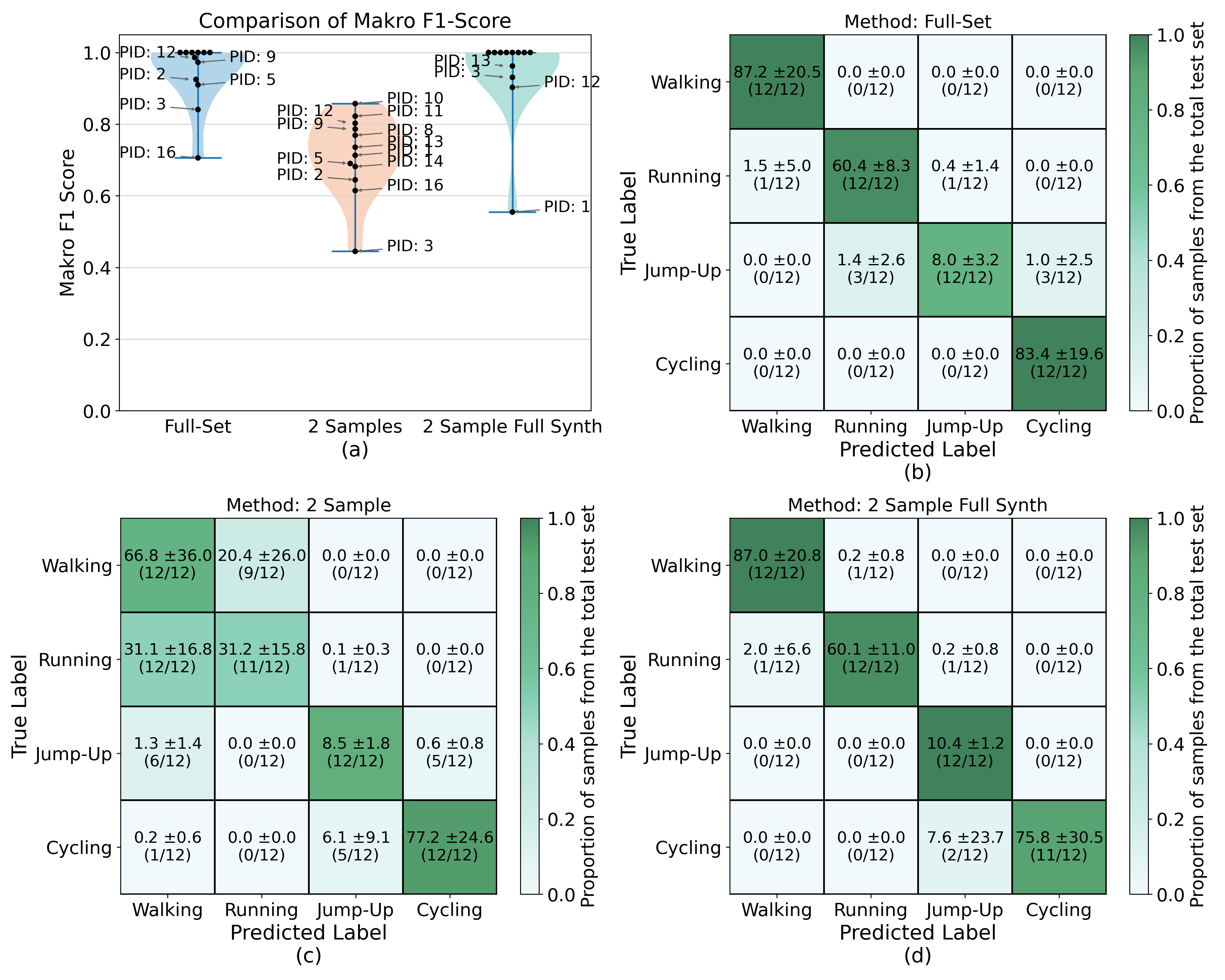}
    \caption{
    Classification results from LOSOCV with and without synthetic sequences in the training set.
    Compared are the two baseline classifier 2 Sample and Full-Set against the classifier which is trained on synthetic and real sequences (2 Sample Full Synth).
    The results were generated only on the real sequences from the hold out test participant. 
    Graph (a) shows the makro F1 score in the form of a swarm plot combined with a violin plot.
    The PIDs are representative for the participant IDs that reached a score value of less than $1.0$.
    (b), (c) and (d) show the confusion matrices from the Full-Set, 2 Sample and 2 Sample Full Synth classifier respectively. 
    It includes the average amount of samples and the standard deviation over all participants. 
    The value in the brackets represents the amount of participants affected by the cell compared to all $12$ participants.
    }
    \label{fig:violinplot}
\end{figure}

The final results from the 2 Sample Full Synth classifier with all synthetic sequences and the baseline classifier are visualized per participant in Figure~\ref{fig:violinplot}.
The synthetic sequences significantly improved the results for almost all participants, compared to the 2 Sample baseline classifier.
Except in four participants, the macro F1-score increased to $1.0$, which means that no sequence in the test set was falsely identified. 
The exceptions were the participants with the IDs $1$, $3$, $12$ and $13$.
Especially the results from PID $1$ were interesting.
It was the only participant where the macro F1-score from the 2 Sample Full Synth classifier decreased compared to the 2 Sample baseline classifier.  
It reduced to to a value of less than $0.6$. \newline
Utilizing the full sample space without synthetic sequences (Full-Set) improved the score value in each participant compared to the 2 Sample classifier.
In six out of the $12$ participants the score improved to a value of $1.0$.
For the remaining six participants the score values range between $0.7$ and slightly below $1.0$.
Worst performance was achieved on PID $16$.
With the 2 Sample Full Synth classifier we were able to improve the score value of this participant to $1.0$. 
The same is applicable for the participants with the IDs $2, 5$ and $9$.
The addition of the synthetic data improved the score value to $1.0$, whereas the Full-Set baseline classifier reached values of less than $1.0$.\newline
With the 2 Sample baseline classifier the two classes Walking and Running were mostly mixed up with each other.
As Figure~\ref{fig:violinplot} (d) indicates, the classifier falsely identified sequences from the Running class as Walking sequences in each participant. 
On the other hand, Walking was misclassified as Running in nine out of $12$ participants.
Interestingly, in one of the twelve participants the model was not able to identify any sequence from the Running class correctly.
By increasing the sample space to the Full-Set, the differentiation between the two classes improved.
The classifier mixed up sequences from the two similar classes in only a single participant.
Conversely, sequences from the minority class were still falsely classified. \newline
Both, the minority class problem as well as the differentiation between the Walking and Running class improved when synthetic data was added. 
The classifier falsely identified sequences from the Running class with Walking as well as the other way around in only a single participant.
On the other hand, no sequence from the minority was falsely identified.
Interestingly though, the 2 Sample Full Synth classifier mistook sequences from Cycling class with sequences from the Jump Up class in two participants.
With one of the two, the classifier was not able to identify a single sequence correctly.
This was not the case with the two baseline classifier.
The classifier using the Full-Set correctly identified each sequence from this class.
In contrast, the 2 Sample baseline classifier mistook sequences from this class either with sequences from the Jump Up or the Walking class. 
This was the case in at least five participants. 
However, it did not occur that all sequences of a participant were incorrectly identified. \newpage

\begin{figure}[H]
    \centering
    \includegraphics[width=0.99\linewidth]{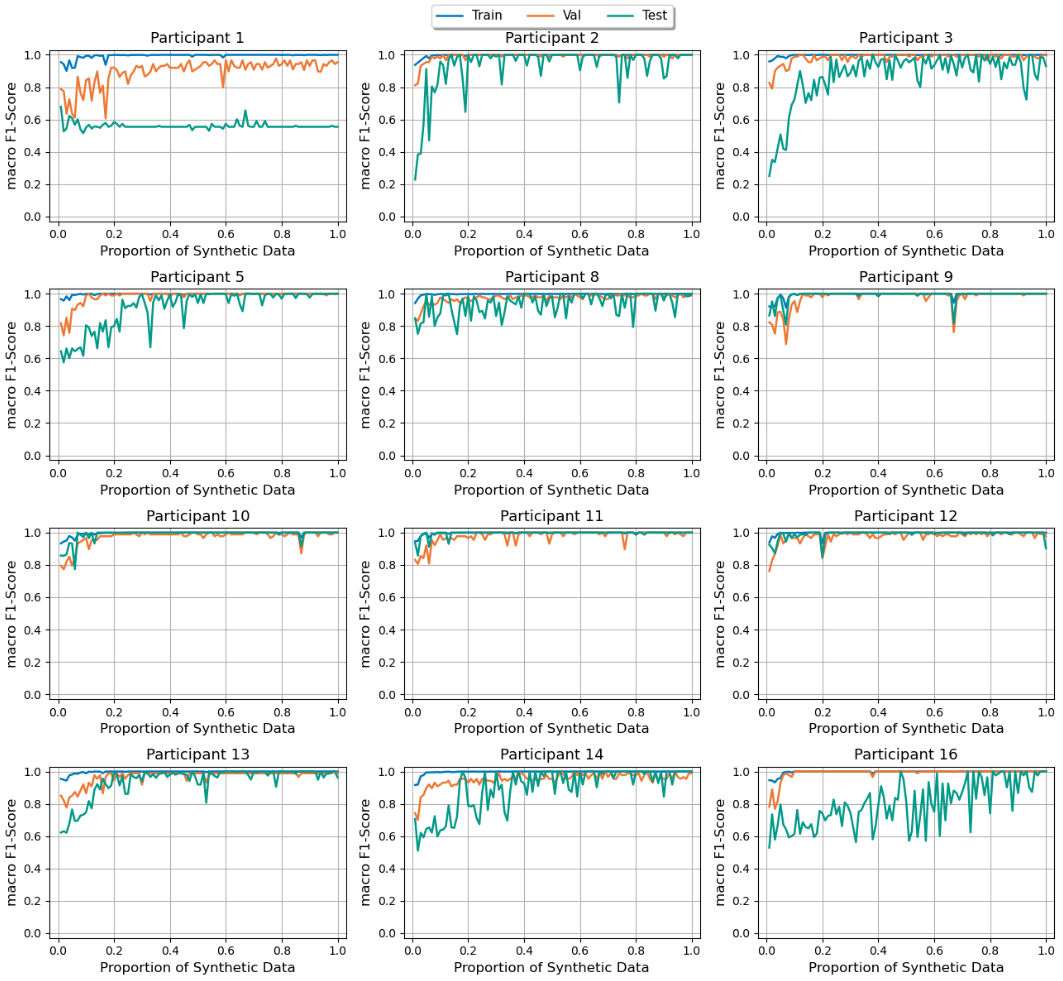}
    \caption{Training results in dependence of the amount of added synthetic data. 
    The entire quantity of the synthetic sequences represents a fraction of $1.0$, which means that $3840$ synthetic sequences were added per class.
    We chose a step size of $1\%$.}
    \label{fig:trainingsverlauf}
\end{figure}

\textbf{Monitoring of the Training Process}\newline
In addition to the evaluation of the full synthetic sample space, we analysed the influence of the amount of added synthetic sequences on the classification task. 
The participant specific results are visualized in Figure~\ref{fig:trainingsverlauf}.
As the performance on the validation set improved gradually for each participant, the performance on the test set fluctuated in some cases, like with PID $14$ and $16$.
Interestingly, by adding only a small amount of synthetic sequences of less than $1500$ per class to the dataset, it is leading to fluctuations in the test score value of about $0.3$ in PID $14$.
By further increasing the amount of synthetic sequences the fluctuations reduced to $0.15$ and less. 
Still, some drops in the score value remained.
They can be seen in each participant.\newline
The performance of the Full-Set baseline classifier was worst on PID $16$.
With only a fraction of three to four percent of the approximately $2600$ real data in the training set, the 2 Sample Full Synth classifier was able to improve the score value to $1.0$ for this participant.
Though, the progression of adding synthetic sequences did not show a clear indication of improvement of the score value up until around $80\%$ of the synthetic data was added.
Even then, there are still some drops of $15$ to $20\%$.\newline
The performance of the 2 Sample Full Synth classifier on PID $1$ was worst throughout all participants.
The score value was less than $0.6$ and remained on the same level even though synthetic sequences were gradually added.
The validation score on the other hand did improve.
It increased by about $0.2$. \newline

\begin{figure}[H]
    \centering
    \includegraphics[width=0.75 \linewidth]{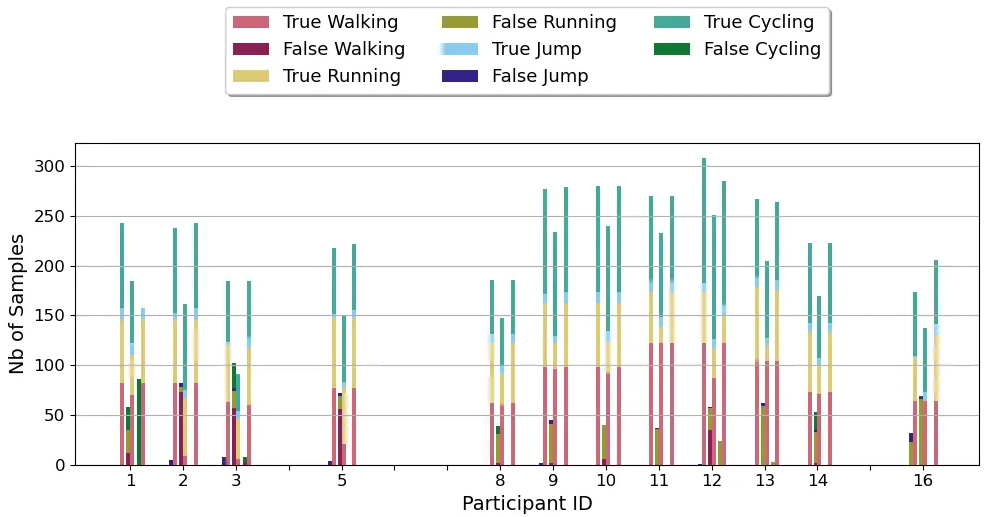}
    \caption{
    Number of true and false identified sequences separated by class and participant.
    Left from the participant ID are the results from the Full-Set baseline model, whereas in the middle the results from the 2 Sample model and on the right side of the ID the results from the IMUDiffusion model are shown.
    }
    \label{fig:participant_barchart}
\end{figure}

\textbf{Participant specific analysis of the classification result} \newline
Figure~\ref{fig:participant_barchart} provides a summary of the participant specific classification results.
It visualizes the number of true and false classified sequences per class and also distinguishes between the three evaluated classifier. 
Six bars are assigned to each participant.
Two belong to the Full-Set, two the 2 Sample and two to the 2 Sample Full Synth classifier.
From left to right, the six bars are assigned to falsely identified sequences from the Full-Set classifier, correctly identified sequences from the Full-Set classifier, falsely identified sequences from the 2 Sample classifier, correctly identified sequences from the 2 Sample classifier, falsely identified sequences from the 2 Sample Full Synth classifier and correctly identified sequences from the 2 Sample Full Synth classifier.
Additionally, each of those six bars is stacked by the class individual results. \newline
Except for PID $1$, the 2 Sample classifier has the worst performance across all participants. 
Responsible for this performance was either the Walking or the Running class.
They accounted for the most misclassified sequences per participant. 
In eight out of the $12$ participants, sequences from the Running class were responsible.
Only in PID $1$ three classes had around the same share of misclassified sequences.
They did belong to the classes Walking, Running and Cycling.
For the 2 Sample Full Synth classifier this was the participant it performed worst on. 
It was not able to identify any sequence from the Cycling class correct, whereas each other class was classified correctly. 
Apart from this participant, the classes per participant that were misclassified remained the same between the 2 Sample and 2 Sample Full Synth classifier.\newline

\newpage

\begin{figure}[H]
    \centering
    \includegraphics[width=0.99 \linewidth]{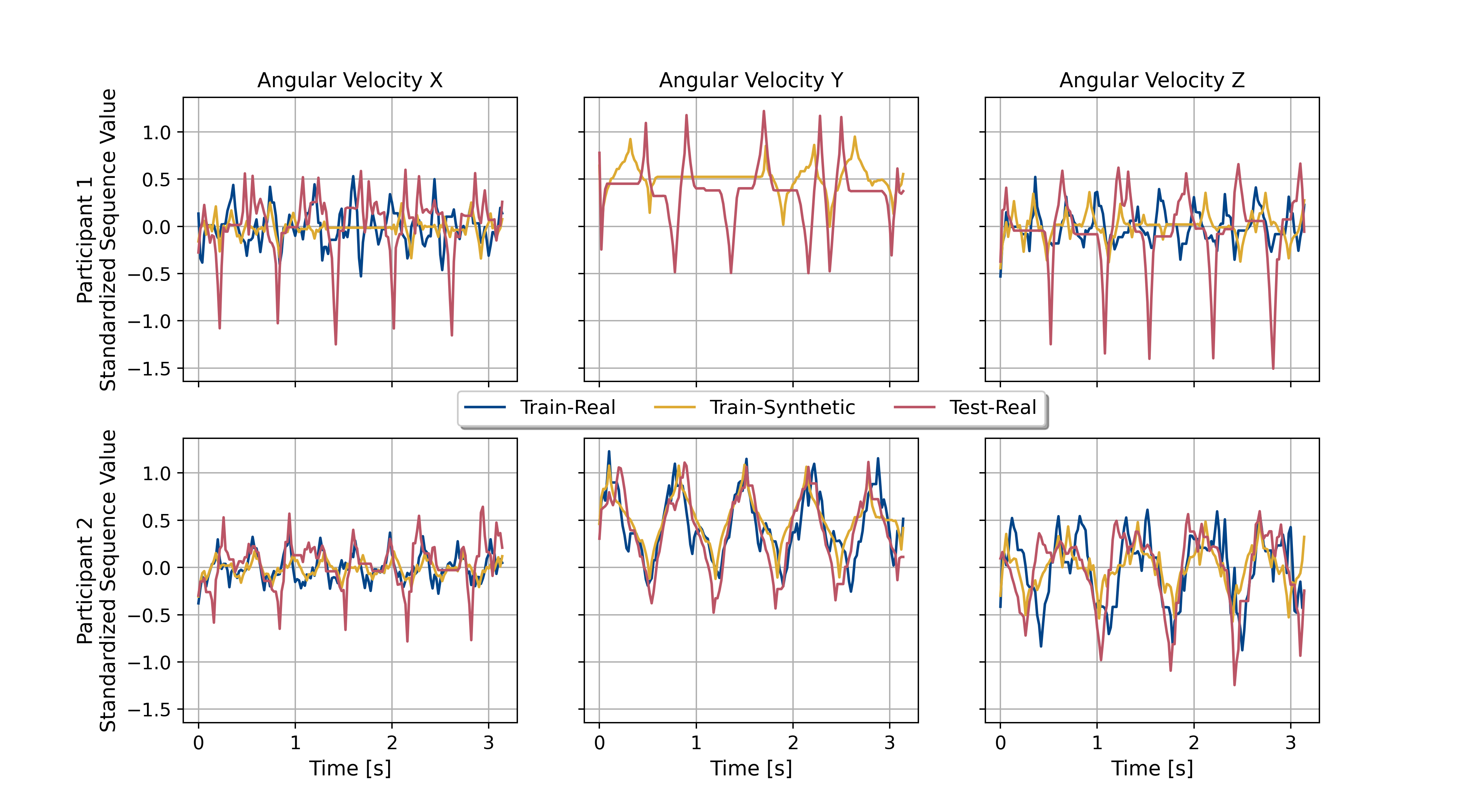}
    \caption{
    An individual cluster analysis was performed on two pre-chosen participants - PID $1$ and $2$.
    The distance metric was DTW and the amount of cluster was set to $20$.
    Visualized is the DBA of one pre-chosen cluster containing the most test sequences.
    For an easier comparison we have calculated the DBA out of the standardized sequences.
    Only the three axes from the Gyroscope are visualized individually for the two participants.
    }
    \label{fig:pid1vspid2}
\end{figure}

In order to analyse the misclassification of the Cycling class of PID $1$ with the 2 Sample Full Synth classifier, we performed a cluster analysis on the real and synthetic training data as well as the hold out test data using kMeans.
The amount of cluster was set to $20$ and it was performed individually on each of the six axis of the IMU.
The results from the gyroscope are visualized in Figure~\ref{fig:pid1vspid2}.
We compared a single cluster from PID $1$ against a single cluster from PID $2$.
The cluster was chosen based on the amount of sequences belonging to the hold out test set.
It was the cluster containing most sequences from this set.
One of the main findings is that the test sequences were assigned to only a single cluster for the angular velocity in y-direction.
No sequence from the real training set was assigned to this cluster.
Apart from this IMU axis, visually we could not find a single cluster with such a high resemblance as between sequences from PID $2$ and sequences from its respective training or synthetic set.\newline

\begin{figure}[H]
    \centering
    \includegraphics[width=0.99 \linewidth]{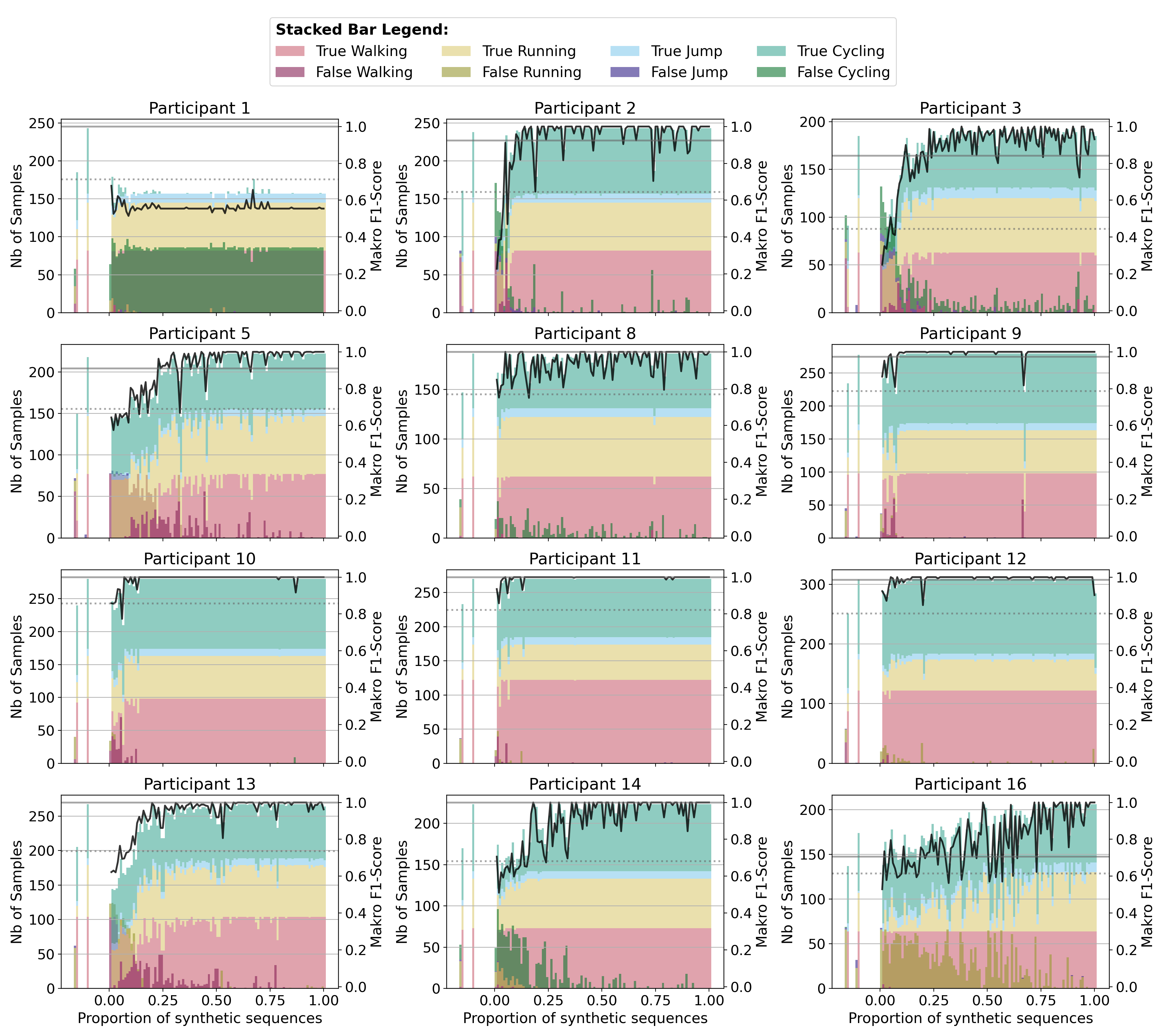}
    \caption{
    Barchart highlighting the process of adding synthetic sequences gradually to the training set with a re-training of the classifier afterwards.
    On the left side, below the zero value within each subgraph, the results from the two baseline classifier are visualized.
    Furthest on the left, the results from the 2 Sample classifier are shown, followed by the ones from the Full-Set. 
    Afterwards, starting from $x=0.01$, the classification results of the 2 Sample Full Synth classifier with the respective amount of added synthetic sequences are visualized. 
    A value of $x=1.0$ indicates the usage of all synthetic sequences within the training set.
    The step size was set to $0.01$.
    }
    \label{fig:gradually_adding_synth_barchart}
\end{figure}

\textbf{Impact of gradually adding synthetic data on the classification result} \newline
Gradually adding synthetic data to the dataset improved the validation score for each participant. 
Though, not each test result improved as expected. 
Figure~\ref{fig:gradually_adding_synth_barchart} visualizes the macro F1-score from each model as well as the amount of sequences per class which were false and correct classified.
Almost independent of the amount of added synthetic sequences, PID $1$ always misclassified the sequences from the Cycling class.
By increasing the sample space for the IMUDiffusion model by less than $200$ synthetic sequences, the same three classes were misclassified as with the 2 Sample model.
By increasing the sample space even further, the amount of classes that led to misclassifications was reduced to the Cycling class solely.\newline
The test score for PID $16$ gradually improves by adding synthetic sequences to the training set whilst still having some drops of the score value.
Misclassified sequences from the Running class were responsible for those drops.
These drops in the score value can be extended to all participants and is even independently of the amount of additionally added synthetic sequences.
Though, even with those drops, the participant individual graphs show some gradual improvement of the score value. 
For the same participants, the classes of the misclassified sequences did not change independently of the amount of added synthetic sequences.
In most cases, the largest share of this was usually held by a single class.

\section{Discussion}
The IMUDiffusion model has the capability of generating high quality human movement patterns. 
In the six axis IMU, the information content varies across the axes, as most of the movement occurred in the sagittal plane.
This way, the main contributors were the x- and y-axis for the acceleration and the y- and z-axis for the angular velocities. 
Based on the visual analysis with UMAP and kMeans, similar sequences across the main contributors were generated.
Though, more impressive is the generation of similar patterns across the minor contributors - the acceleration in z-direction and angular velocity in x-direction.
Still, the visual analysis on the quality of the generated sequences has to improve to get a more generalized feedback.
A standardized metric in this area is required to validate the performance of the model.
We tried to overcome this issue by using the synthetic sequences as additional input for training a neural network which had the task to distinguish between the four classes.
The improvement we achieved with the synthetic sequences confirmed the results from the visual analysis.
We were able to improve on the minority class as well as on the Running class which has high similarities with the Walking class.
Those results demonstrate the opportunity of using a diffusion model across multiple sensor types by adapting the network structure in combination with the scheduler.
It would be interesting to see how effectively different types of sensors interact with each other in the generation process.
One of the main disadvantages of this approach is the required time to generate the synthetic sequences. 
In this paper we used a small dataset of just $22$ sequences per class. 
Still, the required computing time was quite high with $78.4$ hours. \newline
One challenge working with this dataset was the participant with the ID $1$.
We were not able to get the classifier to identify the Cycling class correctly when adding synthetic sequences. 
A cluster analysis revealed an unexpected movement behavior in some axes from the IMU.
Still, the pre-chosen sequences in the 2 Sample dataset showed, that the correct prediction of this class is possible and the Full-Set classifier was even able to improve the score value to $1.0$.
So, it seems, that the synthetisation of sequences from this participant and class was not able to fully grasp the characteristics of the movements. 

\section{Conclusion}
In this paper we introduced IMUDiffusion, a diffusion model for inertial motion capturing systems.
Based on the diffusion model architecture from the Computer Vision domain, we adapted their model to meet the requirements of generating high-quality sequences of human motion.
By using individual scheduler for each sensor type, we were able to generate multivariate sequences across those sensor types.
The quantitative assessment of the synthesised data showed promising results in producing multivariate time series sequences.
Whilst maintaining the characteristics of the signals, it still increased the variability within the dataset. 
Especially, we were able to generate meaningful data despite the scarce amount of available training data. 
Except for one participant, we were able to improve the score value by at least $11\,ppt$. 
Even the baseline model using the fully available dataset was mostly outscored or comparable results were achieved. 
Only a single participant remained a challenge. 
We were not able to correctly classify the Cycling class with this participant.
They were confused with sequences from the Jump Up class. 
Further analysis would be required to clearly identify the challenging sequences as well as to identify the problem with the synthetic sequences from this class. 
Nevertheless, the IMUDiffusion model shows promising results to contribute in the human activity recognition domain and extends the view for time series generation, especially inside a minority class.

\bibliographystyle{unsrt}  
\bibliography{sample}

\section*{Acknowledgements}
The OrthoKI project is funded by the Carl-Zeiss-Stiftung.

\section*{Author contributions statement}
Conceptualization, M.M.; Data curation, H.O.; Formal analysis, H.O. and M.M.; Funding acquisition, M.M.; Investigation, H.O. and M.M.; Methodology, H.O. and M.M.; Project administration, M.M.; Resources, M.M.; Software, H.O. and M.M.; Supervision, M.M.; Validation, H.O. and M.M.; Visualization, H.O.; Writing—original draft, H.O.; Writing—review and editing, M.M. All authors have read and agreed to the published version of the manuscript.

\section*{Competing interests}
The authors declare no competing interests.

\end{document}